# Advancing Household Robotics: Deep Interactive Reinforcement Learning for Efficient Training and Enhanced Performance

[1]Arpita Soni, [2]Sujatha Alla, [3]Suresh Dodda, [4]Hemanth Volikatla

[1]Independent Researcher, Concord, North Carolina, USA, Email: soni.arpita@gmail.com

[2]Old Dominion University, Norfolk, Virginia, USA, Email: salla001@odu.edu

[3]Independent Researcher, Atlanta, Georgia, USA, Email: sureshr.dodda@gmail.com

[4]Independent Researcher, Alpharetta, Georgia, USA, Email: hemanthvolikatla@gmail.com

**Abstract**

The market for domestic robots—made to perform household chore, is growing as these robots relieve people of everyday responsibilities. Domestic robots are generally welcomed for their role in easing human labour, in contrast to industrial robots, which are frequently criticised for displacing human workers. But before these robots can carry out domestic chores, they need to become proficient in a number of minor activities, such as recognizing their surroundings, making decisions, and picking up on human behaviours. Reinforcement learning, or RL, has emerged as a key robotics technology that enables robots to interact with their environment and learn how to optimize their actions in order to maximize rewards. However, the goal of Deep Reinforcement Learning (DeepRL) is to address more complicated, continuous action-state spaces in real-world settings by combining RL with Neural Networks (NNs). The efficacy of DeepRL can be further augmented through interactive feedback, in which a trainer offers real-time guidance to expedite the robot's learning process. Nevertheless, the current methods have drawbacks, namely the transient application of guidance that results in repeated learning under identical conditions. Therefore, we present a novel method to preserve and reuse information and advice via Deep Interactive Reinforcement Learning (DeepIRL)—it utilizes a persistent rule-based system. This method not only expedites the training process but also lessens the number of repetitions that instructors will have to carry out. This study has the potential to advance the development of household robots and improve their effectiveness and efficiency as learners.

## 1. Introduction

Artificial intelligence, the Internet of Things, big data, and other technologies are pivotal for crafting a smart and intelligent healthcare system [1, 2]. In recent years, there have been significant advancements in robot creation and an increased focus on this field[3]–[6]. This achievement is recognized in household settings as well as industrial areas, where robots are gradually replacing humans. The reason for their limited presence in residence environments is primarily the existence of numerous dynamic variables and safety regulations [7]. Furthermore, active human interaction is anticipated to be necessary for agents to respond more effectively, solve problems, and carry out jobs efficiently. Reinforcement Learning (RL) is a technique used by robot controllers to discover the best course of action via trial-and-error interaction with the environment. When an agent is in a particular condition, the policy specifies what it should do. According to the existing policy, the reward function—which is predefined by the environment designer—will provide a reward signal to the agent after it attempts to choose and carry out an action. After completing these processes, the policy will be changed, and this reward will represent the caliber of the acts carried out by the agent. However, the ultimate objective is to discover a policy that optimizes the cumulative total reward.

Deep Reinforcement Learning (DeepRL) is a substitute that utilises the RL framework along with Deep Learning (DL) to provide an approximation for the state value in continuous action-state spaces. Combining the benefits of DL and RL—DeepRL can achieve end-to-end autonomous learning and control using high-dimensional raw environment input data that is mapped to actions in real-world domains. In past, the whole range of scenarios including various learning algorithms, such as Q-learning [8], with exploration methods, such as Greedy, Softmax, have also been examined when applying RL to domestic robots [9]. The outcome demonstrated the huge potential of RL in robot applications. Additionally, DeepRL has shown promise in other domains, including manipulation skills, grasping, and legged locomotion. Nevertheless, the excessive amount of time and resources needed by the agent to obtain satisfactory results remains an unresolved performance issue with the RL and DeepRL algorithms. More computational resources will be required to identify the best course of action in a state space that is larger and more complicated.

While there are many ways to quicken this process, one potential technique that can increase convergence speed and has demonstrated viability is called Interactive Reinforcement Learning (IRL). With IRL, a trainer can assess a learning agent's conduct or offer guidance. In IRL, there are two methods for offering advice. First, rewards shaping—adding rewards to direct the agents—has been found to hasten learning. When compared to the target's context, it could, nevertheless, provide a different scenario. The context with reward shaping has a different solution than the context without reward. Second, policy shaping modifies policies in an effort to make the most of the information obtained from human feedback. In the real world, it can function more efficiently when given highly inconsistent human feedback. Therefore, a Deep Interactive Reinforcement Learning (DeepIRL) model can be employed in continuous space at a faster rate by combining IRL and DeepRL. Compared to the conventional DeepRL approach, the approaches that include external trainers enable the learning agent to archive more rewards, faster learning times, and less mistakes. A novel strategy called persistence rule-based—uses rule-based policy shaping while preserving the value of human counsel for the state's subsequent review—for expediting the learning process. Subsequently, there could be an increase in the agent's performance and a noticeable decrease in the number of interactions during the learning phase. Persistence rule-based approaches are now promising, but they have only been applied to discrete domains such as self-driving vehicle. However, we anticipate that merging persistence rule-based and DeepIRL approaches will enhance the agent's learning capabilities even more and make it easier to create a strategy that works in continuous domains.

The paper is as followsin the next section we will see the background related to our study. In Section 3, the related works are presented. In Section 4, the materials and methods are discussed. In Section 5, the implementation of the framework is presented. In Section 6, the empirical evaluations are presented and we conclude the paper in Section 7 with some conclusion and future works.

**2. Background**

In the Machine Learning (ML) [10]–[21]paradigm known as RL, agents interact with their surroundings to acquire the best possible behaviours. This is a trial-and-error learning process where agents gradually hone their actions to maximise rewards, much like natural learning happens to people and animals. Roots of RL can be found in the behavioural psychology of [22], who popularised the idea of learning through contact with surroundings. With time, RL has developed into a powerful method in computer science that allows agents to maximise their activities to get the greatest results and earn digital rewards. An agent in a RL setup observes the state of its surroundings at each timestep and chooses an action that changes the environment to a new state and rewards it. A Markov Decision Process (MDP), which is composed of states, actions, transitions, and rewards, is used by agents in reinforcement learning (RL) to navigate through an environment and maximise rewards. Reward functions are used by RL to learn instead of pre-established correct actions as in supervised learning. As a function approximator in reinforcement learning (RL), deep learning (DL) is helpful in vast or continuous state spaces. Learning efficacy is increased with the use of an external advisor in techniques such as Inverse Reinforcement Learning (IRL). Furthermore, techniques such as Probabilistic Policy Reuse (PPR) and Ripple-Down Rules (RDR) enable flexible and long-lasting learning in domains such as advanced robotics by improving decision-making in complex and evolving settings. By incorporating DL and RL and supporting it with interactive techniques like IRL and PPR, ML is set to create agents that can solve issues and make complicated decisions in a manner akin to highly evolved cognitive humans. This research represents a significant leap forward in the creation of intelligent systems and has the potential to produce significant advancements in robotics, autonomous systems, and other sectors where autonomous decision-making is essential.

**3. Related Works**

Robots for cleaning houses are self-sufficient mobile devices that can clean windows, floors, lawns, beds, and other surfaces. At home, it has been used for mopping, Ultra Violet (UV) sterilisation, and other uses. Three different forms of robotic navigation technology exists—1) infrared for obstacle avoidance and detection in the 1990s; 2) gyroscope and ground photoelectric sensors with collision-generated maps from 2000 to 2018; 3) LiDAR-based Simultaneous Localisation And Mapping (SLAM) and visual SLAM for enhanced maps and real-time positioning [23]. However, the use of several sensors to control the movement of cleaning robots may be the main emphasis of the upcoming stage. One of the main influences on cleaning robot design, particularly in practical aspects, is technology. The work of [24] outlined the technical difficulties in the design of cleaning robots, including power supply, absolute positioning, and area coverage in uncertain dynamic situations, strong obstacle avoidance, and sensor coverage. [25] have developed an autonomous mobile cleaning robotic system in an obstacle-filled environment by utilising adaptive manufacturing technology and a well-proven control algorithm. A novel glass façade cleaning robot design concept is presented by [26] for effective path-plan modelling and dynamic modelling. [27] suggest a wall-cleaning robot that can increase safety and dependability by adjusting hoover power based on adhesion awareness. [28] developed a domestic random path planning algorithm for cleaning after analysing the performances of three commercial floors. In addition to technology advancement, Human Computer Interaction (HCI) researchers have concurrently examined user adoption of household cleaning robots. For instance, [29] developed a domestic robot ecology paradigm with four temporal stages—re-adoption, acceptance, adaptation, and use/retention after conducting a six-month long field study—on how 30 homes accepted a new cleaning robot. [30] compared seven robots to determine the impact of important technologies and carried out an ethnographic study in nine families to determine the demands of users. Their research points to two design considerations for cleaning robots—a robot's ability to complete a task and its ability to integrate with the user's surroundings and perception.

The adoption of a cleaning robot is further hampered by a lack of confidence in the device and a resistance to making tangible changes at home. According to [31] research, social interactions with robots include playing with a moving robot, identifying your robot, working together to clean, and more. Similar to this, [6, 9] research reveals that by altering who cleans and how, cleaning robots may have an impact on family dynamics. A cleaning robot has the ability to turn cleaning into a family-wide social activity according to [32]. This makes it possible to think of a household cleaning robot as a social service robotic system with the ability to offer greater experiential value. Creating home robots with personalities encourages acceptance and trust from users. Based on personality theories, [33, 34] created three unique robot personas and hypothesised that users would choose robots that mirrored but emphasised their own personalities. Semi-structured interviews were used by [35, 36] to discover that users can interact with robots in a suitable manner if they are aware of the personality traits of the robots. According to [37], children's desire to adopt a home social robot may be influenced by entertainment-related experiences. Social robot experience design raises possible moral dilemmas, such as whether or not humans should maintain emotional bonds with robots. In order to mitigate ethical dilemmas, researchers might reduce the possible constraints of technology by including consumers' real concerns into the design research process. Transparency, predictability, psycho motivational effects, step-by-step information, and explainability are among the suggestions made by [38] to balance experience with ethical design.

## 4. Research and Methodology

### 4.1 Research Design

A major advancement in home automation is the incorporation of intelligent robots into living spaces. It is anticipated that these robots would be able to learn and carry out simple and everyday chores like cooking, cleaning, and organising. Sensing, interpreting, and making decisions are difficult tasks for robots striving for human-like skill in dynamic home situations. To educate robots to emulate human behaviour, inverse reinforcement learning (IRL) is essential for comprehending the incentive structures underlying human behaviour. IRL is being used more and more to train home robots because of its recognition for producing sophisticated behavioural models from observed behaviours. Nevertheless, considerable computational resources are needed to create an ideal strategy. In order to improve learning effectiveness in these kinds of complex situations, researchers propose combining IRL with a persistence rule-based approach, which uses uniform benchmarks to speed up learning in dynamic home environments. By integrating these ideas into the IRL framework, robots can quickly eliminate weak or unsuccessful tasks, focusing their learning efforts on more successful strategies. Nevertheless, most of the work that has already been written about persistent rule-based methods concentrates on discrete domains, where actions and results are well-defined and scarce. The true test is in applying this methodology to continuous domains, where states and actions are continuous and the robot has to make choices in a far more intricate and subtle environment. The persistent rule-based IRL technique is being extended to continuous domains with the goal of bridging this gap. In home robotics, tasks often require an effective balance between precision and adaptability, qualities intrinsic to continuous areas. This would greatly increase the usability of IRL in this context. Therefore, this study is guided by two main research questions. Firstly, can the rule-based persistent IRL technique be applied to continuous representation? In a continuous action-state context, which is more akin to real-world situations, this question asks whether it is theoretically and technically feasible to apply persistent rules. Second, how much can learning in continuous RL scenarios be accelerated by rule-based persistent feedback? The question aims to measure the effect of enduring norms on learning process effectiveness. By establishing a clear correlation between learning speed and persistent rule-based feedback, researchers can confirm the effectiveness of this tactic in improving the performance of domestic robots. The research aims to address these issues by creating a system that combines IRL with permanent rule-based techniques in a continuous setting. The strategy is coming up with a set of general guidelines that can be used in a home under different conditions. Next, by evaluating the robot's ability to learn and perform tasks, the framework must be verified and put to the test via intensive simulations and real-world experiments. Flexibility and scalability will also be examined in this research.

### 4.2 Research Methodology

In order to illustrate the significance of persistent feedback mechanisms in intelligent systems, this study examines how DL, RL, DeepRL, IRL, and other techniques are used in household robots. While DL uses multi-layered neural networks to analyse data, RL uses trial and error to train agents how to interact with their environment and maximise rewards. With the combination of DeepRL, agents can handle high-dimensional inputs and solve intricate issues, demonstrating major breakthroughs in fields such as robotics and gaming. Human feedback is included into IRL to improve learning speed and adaptation in changing home circumstances. A systematic rule-based method in IRL called persistent feedback shortens the time and energy required for agents to become proficient at activities in unpredictably changing environments. First testing ground to put these concepts into practice is the Cart Pole scenario from AI gym environments, which replicates the kind of balancing issues faced by household robots. Further complexity is introduced by using simulation tools such as Webots or CoppeliaSim to generate a home-like environment with obstacles like as walls, furniture, and more (see Fig. 1). Inaccurate or overly negative feedback might have the opposite impact and lead to inefficiencies or the wrong learning routes, even if timely and accurate feedback can significantly boost learning. Experiments that measure agent performance under three different feedback conditions—no interaction, non-persistent interaction, and persistent rule-based interaction—help us understand how these approaches affect learning effectiveness and the practical performance of domestic robots.

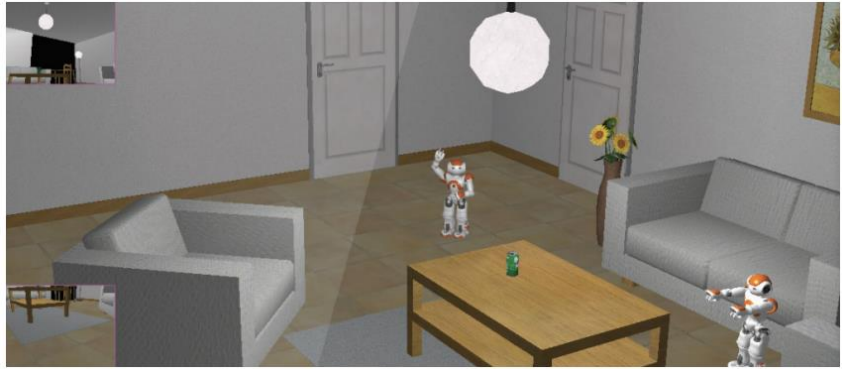

Fig. 1. A Webots environment example

**5. Implementation**

The study will employ Python, which is renowned for its versatility and extensibility and can be applied to a wide range of tasks. Its ease of understanding and abundance of libraries make it a perfect choice for developing complex algorithms in the field of reinforcement learning. DeepRL will operate in the AI gym environment, which has received recognition from the RL community for its durability. The primary illustration of a continuous state environment will be the inverted pendulum game, also known as Cart Pole. This game tests the agent's ability to react dynamically to changes in the environment by having the player manoeuvre the pole left and right to maintain its balance on the cart. There will be two main steps in the review process. First, this environment will be used to train simulation agents. These agents will go through a series of tests to determine the number of runs necessary for the findings to converge, taking into account their prior knowledge of the environment. One agent will be chosen for additional testing if they show consistently positive results after three test runs. The second phase will concentrate on the simulation adviser, whose job it is to guide a newly recruited agent who has never been in an area before. The efficacy of the advisor will be evaluated according to the frequency and accuracy of the advice provided (refer to Table 1). For example, a 70% accurate advisor should give accurate advise 70% of the time, with wrong advice likely to be given in the other cases. This divergence enables a more advanced understanding of how advise frequency and accuracy affect the agent's learning process. Three agents—realistic, optimistic, and pessimistic will be created to provide a thorough evaluation of the system. In order to provide a wide range of conditions for evaluating the efficacy of the PPR model in action selection, these agents will represent varying degrees of advisory accuracy and frequency. PPR functions on the principle that initial adherence to previously advised courses of action is preferred, with a progressive reduction in this dependence over time to facilitate the shift to more independent decision-making based on the $\varepsilon$-greedy action selection policy as shown in Fig. 2. Six different agent types will be included in the experimental framework to investigate the aspects of persistence in the learning process (refer to Table 2). These agents will be carefully documented for clarity and convenience of reference throughout the study. They will be categorised based on the presence or absence of the persistence model and different levels of advisory engagement. The study will tackle the problem of implementing rule-based models in continuous state spaces, moving from theory to practice (refer to Fig. 3). Continuous environments are more complex than discrete settings because there could be an unlimited number of states, as opposed to finite and manageable states in discrete environments. For instance, the CartPole environment includes several continuous features, such as the position and speed of the cart as well as the angle and rate of rotation of the pole. Feature extraction will be used to reduce the complexity of these continuous variables into a manageable set, which will make it easier to apply the RDR methodology to the construction of a rule-based model. The final product of the research will be the creation of a simulated home environment with the aid of programmes like CoppeliaSim and Webots. The robot will be tasked with navigating from an initial to a target point while dodging various hazards in this environment, which is intended to replicate the complexities of a household. For the algorithms and techniques that have been perfected in the earlier stages of the research, this environment will act as the ultimate testing ground, offering a thorough understanding of their applicability and efficacy in real-world circumstances. The knowledge gathered from these tests should greatly advance robotics by improving the efficiency and autonomy of robots in residential environments.

Table 1. For the experiments, three virtual users were created

| Agent | Frequency Accuracy |
|---|---|
| Pessimistic Advisor | 23.658% |
| Real Advisor | 47.316% |
| Optimistic Advisor | 100% |

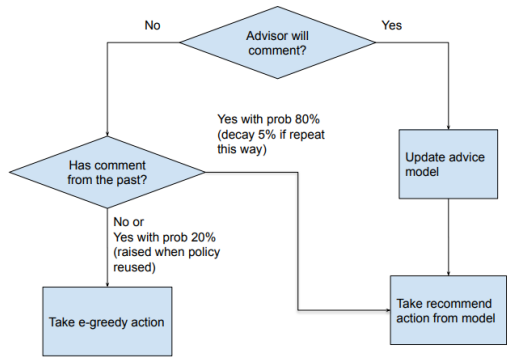

Fig. 2. PPR for an IRL agent that has an 80% chance of using previously given guidance

Table 2. Combinations of simulation advisors for ongoing agent testing

| Agent | Short Name |
|---|---|
| Non Persistence Pessimistic Advisor | NPP |
| Non Persistence Real Advisor | NRP |
| Non Persistence Optimistic Advisor | NPO |
| Persistence Pessimistic Advisor | PP |
| Persistence Real Advisor | PR |
| Persistence Optimistic Advisor | PO |

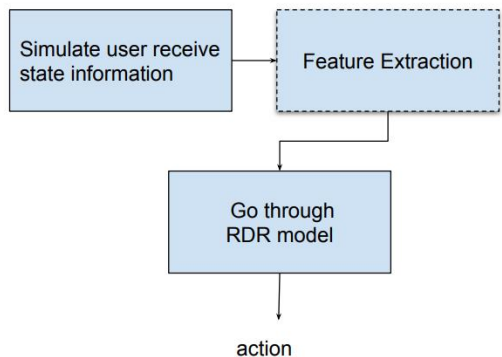

Fig. 3. The rule-based approach's flow

## 6. Empirical Evaluations

The evaluation of intelligent robotic systems in home settings depends on the intricate fusion of multiple ML techniques, most notably IRL and DeepRL, with continuous feedback mechanisms. This investigation uses the CartPole scenario as a testbed for continuous action-space modelling, focusing on the implementation and effectiveness of these techniques within the Python AI gym environment. The creation of a neural network model specifically designed for action selection in the CartPole environment—a popular benchmark in studies on RL—is essential to this evaluation. The goal of the model's design is to give the agent the ability to balance the pole on the cart for a longer period of time, imitating the dynamic balancing duties that are performed by household robots. The model's performance is evaluated against the average rewards accumulated over a number of episodes. Important parameters, including the initial epsilon value, epsilon decay rate, learning rate, and discount factor, are carefully tuned to optimise the learning process. The first trials in the CartPole environment show how the model can adjust and get better at what it does (refer to Fig. 4). It shows how rewards build up gradually and how learning outcomes stabilise after a given amount of episodes. An interactive feedback advisor is included to improve the efficiency and further refine the learning process by giving the agent precise, real-time recommendations. This advisor is intended to accelerate the agent's learning curve by providing accurate and timely guidance. It is designed to be 100% accurate and responsive. Comparing agent performance with and without advisor feedback highlights the important influence of interactive advice on learning rate and overall efficacy (refer to Fig. 5). The assessment advances to include a consistent rule-based methodology, building on the basis of interactive feedback. By remembering and using learnt rules consistently during the learning phase, this approach seeks to expedite the agent's decision-making process. Through less reliance on trial-and-error learning, this method aims to reduce the amount of time and computational resources needed for the agent to operate at optimal levels. This approach is tested using several types of agents, all of which have different levels of accuracy and availability of input. These agents are thoroughly trained in the CartPole environment, where every action they take is monitored and assessed. The goal of the project is to improve reinforcement learning (RL) in home robots by simulating real-world home issues with simulation tools such as Webots or CoppeliaSim. By combining permanent rules with interactive feedback based on experimental findings, it seeks to optimise reinforcement learning techniques. This strategy entails building a neural network model, adding interactive input, using persistent rule-based techniques, and running testing in a home environment simulation. This methodology offers a strong foundation for assessing comprehensive reinforcement learning techniques, providing insights into how to best optimise learning processes for intelligent robotic systems, and eventually advancing home automation technologies and their use in routine household chores.

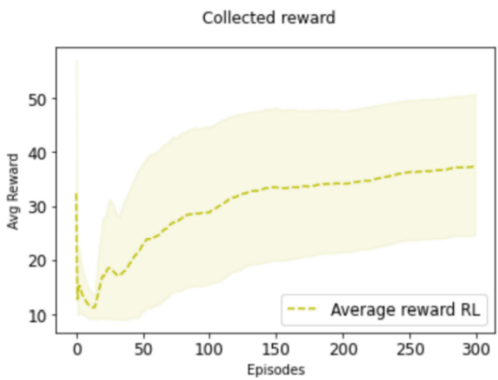

Fig. 4. The CarlPole environment was used to create 300 DeepRL episodes

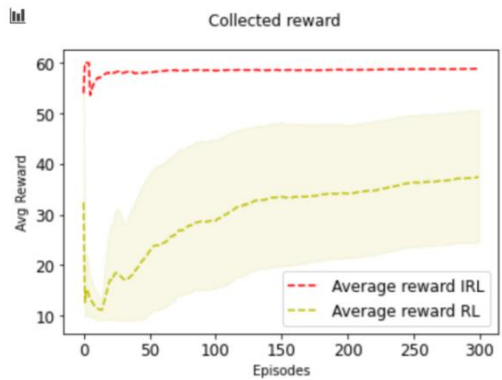

Fig. 5. CarlPole environment was used to create 300 episodes for IRL as compared to RL

## 7. Conclusion and Future Works

In this study, we conducted a comprehensive literature review to establish the foundational knowledge required for our research, and subsequently devised a methodological approach divided into five distinct sprints. These sprints encompassed the development of a built-in environment, the implementation of interactive feedback mechanisms, the utilization of persistent methods, the application of rule-based approaches, and the creation of an environment tailored for home robots. Notably, we successfully created an integrated DeepRL environment and made significant progress in the implementation of an advice interaction system, marking the completion of the second sprint. Recent findings indicate promising advancements, particularly in the realm of IRL algorithms, where the learning pace demonstrated noticeable improvements, especially when compared to standard RL techniques. However, our future endeavors necessitate further experimentation, particularly with real agents and the inclusion of both optimistic and pessimistic agents to comprehensively evaluate algorithm performance. Moving forward, our focus will involve the continuation and completion of the remaining sprints outlined in our research strategy. Additionally, we will conduct additional trials and experiments to validate our findings and refine our methodologies. By doing so, we aim to contribute to the advancement of research in the intersection of RL and interactive systems, with potential implications for various fields, including robotics and AI.